\documentclass[sigconf,anonymous=false,nonacm=true]{acmart}

\AtBeginDocument{%
  }

\setcopyright{cc}
\setcctype[4.0]{by}

\usepackage[utf8]{inputenc}
\usepackage{float}
\usepackage{graphicx}
\usepackage{booktabs,multirow,multicol}
\usepackage{subfig}
\usepackage{caption}
\usepackage{amsmath}
\usepackage{xspace}
\usepackage{xcolor}

\newcommand{\h}{\mathbf{h}\xspace}
\newcommand{\W}{\mathbf{W}\xspace}

\newcommand{\K}{\mathbf{K}\xspace}
\newcommand{\A}{\mathbf{A}\xspace}
\newcommand{\Q}{\mathbf{Q}\xspace}
\newcommand{\V}{\mathbf{V}\xspace}

\newcommand{\CG}{\mathcal{G}\xspace}

\newcommand{\CV}{\mathcal{V}\xspace}
\newcommand{\CE}{\mathcal{E}\xspace}

\newcommand{\CX}{\boldsymbol{X}\xspace}

\newtheorem{defn}{\textbf{Definition}}

\newtheorem{prob}{\textbf{Problem}}

\newcommand{\namemodel}{NeuroSteiner\xspace}
\newcommand{\REST}{REST\xspace}

\definecolor{darkgreen}{rgb}{0.0, 0.5, 0.0}
\definecolor{brightpink}{rgb}{1.0, 0.0, 0.5}
\definecolor{brilliantrose}{rgb}{1.0, 0.33, 0.64}

\begin{document}

\title{NeuroSteiner: A Graph Transformer for Wirelength Estimation}

\author{Sahil Manchanda}
\affiliation{\institution{Indian Institute of Technology Delhi}\country{}}
\authornote{Work completed during an internship at Qualcomm AI Research.\\Qualcomm AI Research is an initiative of Qualcomm Technologies, Inc.}
\email{sahilm1992@gmail.com}
\author{Dana Kianfar}
\affiliation{\institution{Qualcomm AI Research}\country{}}
\email{dkianfar@qti.qualcomm.com}
\author{Markus Peschl}
\affiliation{\institution{Qualcomm AI Research}\country{}}
\email{mpeschl@qti.qualcomm.com}
\author{Romain Lepert}
\affiliation{\institution{Qualcomm AI Research}\country{}}
\email{romain@qti.qualcomm.com}
\author{Michaël Defferrard}
\orcid{0000-0002-6028-9024}
\affiliation{\institution{Qualcomm AI Research}\country{}}
\email{mdeff@qti.qualcomm.com}

\begin{abstract}
    A core objective of physical design is to minimize wirelength (WL) when placing chip components on a canvas.
    Computing the minimal WL of a placement requires finding rectilinear Steiner minimum trees (RSMTs), an NP-hard problem.
    We propose NeuroSteiner, a neural model that distills GeoSteiner, an optimal RSMT solver, to navigate the cost--accuracy frontier of WL estimation.
    NeuroSteiner is trained on synthesized nets labeled by GeoSteiner, alleviating the need to train on real chip designs.
    Moreover, NeuroSteiner's differentiability allows to place by minimizing WL through gradient descent.
    On ISPD 2005 and 2019, NeuroSteiner can obtain 0.3\% WL error while being 60\% faster than GeoSteiner, or 0.2\% and 30\%.
\end{abstract}

\maketitle

\begin{figure*}
    \includegraphics[width=\linewidth]{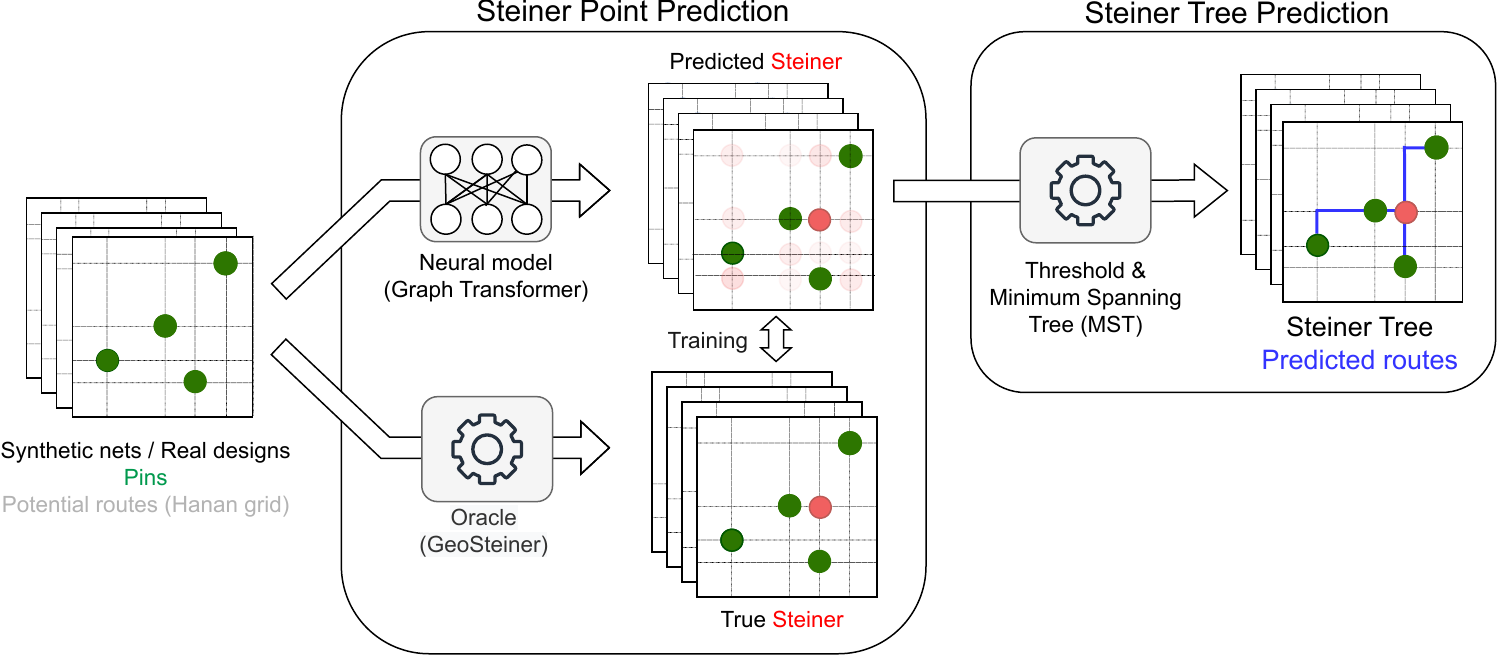}
    \caption{
        \namemodel.
        Rectilinear Steiner Minimum Trees (RSMTs), hence wirelength (WL), are predicted in two steps: (1) determine Steiner points, then (2) find its Minimum Spanning Tree (MST).
        Our neural model predicts the probability that each node on the Hanan grid is a Steiner point.
        The model is trained by distilling an oracle: synthesized nets (or real designs) are labelled by GeoSteiner, an optimal RSMT solver.
    }\label{fig:arch}\label{fig:hanan}
\end{figure*}

\section{Introduction}\label{sec:intro}

Optimizing the placement of circuit elements, e.g., standard cells, in integrated circuits is a critical step in early phases of Electronic Design Automation (EDA)~\cite{chuflute,shahookar1991vlsi,kahng2011vlsi}.
A placer optimizes the location of circuit elements based on power, performance, and area (PPA) objectives. 
One crucial objective is \textit{wirelength} (WL), which is a common proxy for power consumption~\cite{mirhoseini2021graph,shahookar1991vlsi,caldwell1998wirelength}.
In principle, a WL function estimates the expected total length of wires needed to connect all nets in a routed netlist.
Due to the iterative nature of (gradient-based) placement optimization, any desirable WL function needs to be fast to evaluate, accurate, and differentiable.


    %

Among popular WL estimation methods~\cite{shahookar1991vlsi, kahng2011vlsi}, the Rectilinear Steiner Minimum Tree (RSMT)~\cite{kahng2011vlsi} is known to be an accurate estimator of the true routed WL~\cite{jarrod2007}.
Specifically, solving for an RSMT consists of adding nodes—the Steiner points—to a graph such that the length of its spanning tree is minimal with respect to the L1 (Manhattan) metric.
While GeoSteiner~\cite{juhl2018geosteiner}, an algorithm that enumerates Steiner trees, is optimal, its runtime is exponential in the size of the problem (i.e., the net degree) since finding the optimal set of Steiner points for an arbitrary point set in a plane is NP-hard~\cite{zhang2016rectilinear}. 

Several works have used efficient and heuristic approximations to the RSMT with desirable empirical performance.
The Minimum Spanning Tree (MST) can be viewed as the simplest approximation because it assumes no Steiner points and incurs a runtime cost of order $\mathcal{O}(d\log{}d)$ for a net of degree $d$.
However, it overestimates WL by $4\%$ on average~\cite{wong2008scalable}.
Similarly, FLUTE~\cite{wong2008scalable} uses a look-up table to quickly approximate WL without solving for any Steiner points, but is non-differentiable with respect to pin locations.
Bi1S~\cite{kahng1992new} proposes a batched-iterative approach to solving the RSMT problem.
However, its performance is subpar on low-degree nets which constitute the majority of real-world netlists~\cite{chuflute}.
Another method SFP~\cite{sfp} leverages CPU/GPU parallelization to accelerate the RSMT construction.
Despite its high throughput, its solution quality rapidly degrades with net degree.

While these algorithms were hand-engineered to trade off accuracy for runtime, machine learning (ML) offers an automatic way to navigate this trade-off through the distillation of an optimal algorithm.
Moreover, fine-tuning the neural model allows for tailoring it to the real data distribution (not the worst case).
Accordingly, there has been a surge in developing ML approximations to combinatorial optimization problems on graphs~\cite{kool2018attention,khalil2017learning,bengio2021machine,gasse2019exact}.
Finally, while progress in hardware benefits all workloads, the current emphasis on ML accelerates these workloads faster: GPUs are being optimized for calculations used by neural models.



To our knowledge, \REST~\cite{liu2021rest} is the only method that approached the RSMT problem with ML. \REST uses a reinforcement learning (RL) approach, proposing an auto-regressive model that sequentially adds edges to a tree to predict an RSMT.
However, \REST learns multiple models that are specialized for nets of particular degrees (number of pins in a net) instead of a single model for all input sizes.
Furthermore, due to its auto-regressive nature, it generates solutions in a step-by-step sequential fashion which can increase runtime.
Overall, this can lead to long runtime when predicting the wirelength of a whole netlist, consisting of a large variety of net degrees.


\subsection{Contributions}\label{sec:contribution}

\paragraph{\textit{One-shot} supervised binary node classification task}
We formulate the prediction of Steiner points as a \textit{supervised} binary node-classification task where each intersecting point in the Hanan grid (see Figure~\ref{fig:arch}) is considered as a node in a graph.
\namemodel leverages a Graph Transformer~\cite{rampavsek2022recipe} through training on labeled data from GeoSteiner~\cite{juhl2018geosteiner}. The prediction of Steiner points is done in a \textit{one-shot} fashion, facilitating the utilization of GPU parallelization. In contrast to REST, we only predict Steiner points and not a tree. Thus, we propose a hybrid method, where the neural network focuses on approximating the NP-hard problem of finding Steiner point(s), while the MST calculation can be done in polynomial time.
\paragraph{Training on infinitely available synthetic nets}
As obtaining large amounts of industrial chip designs is expensive, we train \namemodel on synthetically-generated labelled data that is cheaper to obtain. We show that the performance of \namemodel can then be improved by fine-tuning on industrial datasets to adapt to the real data distribution.

\paragraph{Evaluation on real-world benchmarks}
Through extensive experiments on chip design benchmarks from ISPD2005 and ISPD2019, we establish that \textsc{NeuroSteiner} \textbf{(1)} constructs an RSMT estimate with an error of about $0.3\%$ when compared to the optimal solution and \textbf{(2)} predicts RSMTs for the benchmarks at 0.3 milliseconds per net on average.

%

\section{Methodology}
Formally, our goal is to solve the following problem:

\begin{prob}[Rectilinear Steiner Minimum Tree]\label{prob:RSMT}
    Given a set of points $\CV_P \in \mathbb{R}^2$,
    construct a rectilinear minimum spanning tree connecting a set of  points $\CV \in \mathbb{R}^2$, with $\CV \supseteq \CV_P$.
\end{prob}

In the above definition, the newly introduced points $\CV \setminus \CV_P$ are called Steiner points.
Constructing the Rectilinear Steiner Mimimum Tree (RSMT) for a set of points is known to be NP-complete~\cite{hanan1966steiner,chuflute}.
However, one can show that for any set $\CV_P$, there always exists a set of optimal Steiner points on the Hanan grid \cite{hanan1966steiner}, which is the union of nodes resulting from intersecting all horizontal and vertical lines passing through each point $v \in \CV_P$.
For example, Fig.~\ref{fig:arch} shows green nodes which depict the set of pins on the 2D plane and the lines crossing the green points represent the Hanan grid.
Let $\CV_H$ denote Hanan points, i.e., the set of intersecting points apart from the pins.
We model the RSMT problem as a binary classification problem of the points in $\CV_H$.

\begin{prob}
    Given a set of pins $\CV_P \subseteq \mathbb{R}^2$ and corresponding Hanan points $\CV_H \subseteq \mathbb{R}^2$, learn the parameters of a neural model for predicting Steiner points in $\CV_H$.
\end{prob}

Fig.~\ref{fig:arch} shows the prediction and training flow of \namemodel.

\subsection{Hanan grid graph}

Given a set of pin points $\CV_P$, we first identify its Hanan points $\CV_H$.
To represent relationships among various points in a Hanan grid, we represent the grid using a graph.
Let the graph $\CG=(\CV,\CE)$ be the graph consisting of the node set $\CV =\CV_H \bigcup \CV_P $ and the set of edges $\CE$ consisting of the Hanan grid.
Formally, let $\CE = \{  {e_{uv} = (u, v) \ | \ v \in \mathcal{N}(u)}  \} $ be defined as the set of neighboring edges, where neighbors are defined as follows:

\begin{defn}[Neighborhood of a node]
    In a graph $\CG$, a node $v$ is said to be a neighbor of node $u$, i.e., $v \in \mathcal{N}(u)$, if and only if one of the following conditions hold $\forall w \in \CV \setminus \{u,v\}$ and  $\forall \alpha, \beta \in [0,1]$:
    \begin{align}
        v_x = u_x \ \text{ and }& \ w_x \neq \alpha \cdot v_x + (1 - \alpha) \cdot u_x, \label{eq:nbrx} \\
        v_y = u_y \ \text{ and }& \ w_y \neq \beta \cdot v_y + (1 - \beta) \cdot u_y, \label{eq:nbry}
    \end{align}
    where $v_x$ refers to the $x$ coordinate of node $v$ and $v_y$ refers to its $y$ coordinate.
\end{defn}

In simple terms, eq.~\ref{eq:nbrx} holds if two nodes $u$ and $v$ lie on the same horizontal line and no other node lies horizontally between them. Eq.\ref{eq:nbry} can be viewed analogously for the vertical dimension.

\subsection{Node and edge features}
Apart from $\CG$, we define a set of node and edge features containing task-relevant information to be used as input to our neural network.
As \textit{node features}, we make use of positional information and a node pin indicator variable.
Formally, the input features of a node $v$ are defined as
\begin{equation}\label{eq:gnn_initialize}
    \h_v^0 = [v_x, v_y, \textit{I}(v)].
\end{equation}
In the above equation, $v_x$ and $v_y$ denote the $x$ and $y$ coordinates of a node $v$, and $I(v)$ is an indicator function that returns $1$ if $v \in \CV_P$ and $0$ otherwise.
The coordinates $v_x$ and $v_y$ serve as positional information, which assists our model to recognize spatial dependencies between different nodes~\cite{rampavsek2022recipe}.
Since our Graph Transformer model employs message-passing between different nodes to construct node embeddings, we employ the pin indicator $I(v)$ as a mechanism for messages to differentiate pin nodes $v\in \CV_P$ from candidate Steiner nodes $v \in \CV_H$.
Let $\CX^0 = \left[\h_v^0\right]_{v\in\CV} \in \mathbb{R}^{\lvert \CV \rvert \times 3}$ denote the input feature matrix of nodes belonging to the graph $\CG$.

For the \textit{edge features} of an edge ${e}_{uv} \in \CE$, we define
\begin{equation}
    \mathbf{e}_{uv} = [  u_x - v_x  ,  u_y - v_y   ]
\end{equation}
to capture the displacement between two neighboring nodes on the Hanan grid.

\subsection{Message-passing and Graph Transformers}

From the input node features, we aim to generate a richer representation that encodes local and global structural information. Since classification of a node as a Steiner point is influenced by nodes beyond its local neighborhood, we need a model that can capture long-range dependencies. While, theoretically, message passing neural networks (MPNNs)~\cite{NEURIPS2021_f1c15925,rampavsek2022recipe} are able to learn global interactions, they require several rounds of message-passing at the risk of oversquashing and oversmoothing~\cite{gnn_exp,morris2019weisfeiler,NEURIPS2021_f1c15925}. Instead, we opt for Graph Transformers, combining  MPNNs with a global attention mechanism~\cite{vaswani2017attention}. This facilitates capturing interactions at both local and global levels ~\cite{rampavsek2022recipe}.

More precisely, we use the GraphGPS~\cite{rampavsek2022recipe} Graph Transformer.
In GraphGPS, one can use any type of MPNN to aggregate information from the local neighborhood of a target node. Here, we use GINE~\cite{xu2018powerful}, updating the node embeddings $\mathbf{h}_v^{(l)}$ at layer $l$ by
\begin{equation}\label{eq:local_mpnn}
    \mathbf{h}_v^{(l+1)} = \operatorname{MLP}^{(l)}\left( \mathbf{h}_v^{(l)}  + \sum_{u\in\mathcal{N}(v)}\phi\left(\mathbf{h}_v^{(l)} + \mathbf{e}_{vu} \right) \right),
\end{equation}
where $\phi: \mathbb{R} \rightarrow \mathbb{R}$ is an activation function, in our case chosen to be the ReLU activation $\phi(z) = \max\{0, z\}$. The edge features $\mathbf{e}_{uv}$ are projected to the same dimension as $\mathbf{h}_v$ using an MLP layer. At a given layer $l$, the above equation aggregates features of neighboring nodes and features of their associated edges.
This information is used to generate local embeddings of nodes at layer $l+1$ represented by $\CX_{\text{loc}}^{(l+1)}=\left[\h_v^{(l+1)}\right]_{v\in\CV}$.

For the global attention scheme, we process the node embeddings at layer $l$ as follows
\begin{equation}\label{eq:glob_attn}
    \quad \quad  \Q{=}\CX^{(l)} \W_\Q, \quad \; \; \; \K{=}\CX^{(l)} \W_{\mathbf{K}}, \quad \V{=}\CX^{(l)} \W_\V , \;
\end{equation}
\begin{equation}
    \CX^{(l+1)}_{\text{glob}} = \operatorname{softmax}(\A) \V \quad \text{where} \quad \A {=}\frac{\Q \K^{T}}{\sqrt{d}}.
\end{equation}
The input features $\CX^{(l)}$ at layer $l$ are projected by three learnable matrices $\W^l_\Q \in \mathbb{R}^{{d\times d}}$,  $\W^l_\K \in \mathbb{R}^{{d\times d}}$, and $\W^l_\V \in \mathbb{R}^{{d\times d}}$ to the corresponding
representations $\Q, \K, \V$
and the attention matrix $\A \in \mathbb{R}^{|\CV| \times |\CV|}$ captures the attention weight between all pairs of nodes.




From the local and global node representations at layer $l$, their representation at the next layer are given by
\begin{equation}\label{eq:loc_glob}
    \CX^{(l+1)} = \operatorname{MLP}\left( \CX_{\text{loc}}^{(l+1)} + \CX_{\text{glob}}^{(l+1)} \right).
\end{equation}
With GraphGPS, we perform $L$ rounds of local message-passing and global attention to obtain the final representation $\CX^{(L)}$ of all nodes in $\CG$.
The more the layers, the more complex relationships GraphGPS can capture.
Finally, embeddings pass through an MLP to obtain the logits
\begin{equation}
    \hat{\mathbf{Y}} = \operatorname{MLP}_{\text{out}}\left(\mathbf{X}^{(L)}\right) \in \mathbb{R}^{|\CV|}.
\end{equation}
These logits represent the probability that a node is a Steiner point.

\subsection{Training}

The parameters of the model are learned by minimizing an objective that encourages the GraphGPS neural model to predict the true labels $\mathbf{Y}$.
The binary cross-entropy objective is calculated as
\begin{equation}
    -\frac{1}{N} \sum_{i=1}^N \mathbf{Y}_i \cdot \log \hat{\mathbf{Y}_i} + \left(1-\mathbf{Y}_i\right) \cdot \log (1-\hat{\mathbf{Y}_i}),
\end{equation}
$\mathbf{Y} \in \{0,1\}^N$ are the true labels, $\hat{\mathbf{Y}} \in \mathbb{R}^N$ are the predicted logits, and $N$ is the number of Steiner candidates in the Hanan grid. We note that this loss models each nodes as an independent Bernoulli random variable.

\subsection{Inference}

After training, for an unseen problem instance we (1) predict the probability of each node in the Hanan grid to be a Steiner point with our Graph Transformer, (2) classify Steiner points by thresholding\footnote{Nodes whose score $\hat{\mathbf{Y}}_i$ are greater than the threshold are classified as Steiner points.}, and (3) obtain the RSMT by finding the Minimum Spanning Tree (MST) from the original pins and predicted Steiner points.
Wirelength is computed by the sum length of edges in the RSMT.


\section{Experiments}

In this section, we benchmark the performance of \namemodel against several existing methods on real chip designs.
We aim to answer the following questions:
\begin{description}
    \item[Sec.~\ref{sec:results}] How does \namemodel perform in terms of \textit{wirelength (WL) estimation error} (accuracy) and \textit{runtime} (cost)?
    \item[Sec.~\ref{sec:synthetic-real}] What is the impact of training on synthetic or real nets?
    \item[Sec.~\ref{sec:scaling}] How to explore the cost--accuracy frontier with model capacity?
\end{description}

\subsection{Setup}

To get a sense of the cost--accuracy frontier, we evaluate \namemodel at two capacities: \textit{small} with $L=10$ layers and \textit{large} with $L=20$.
At both capacities, our GraphGPS model features hidden representations of size $d=64$ and one attention head.
On our setup, our large model is trained with a batch size of 12 while the small model can afford 16.
We train it with the Adam optimizer~\cite{kingma2014adam} with a learning rate of $10^{-4}$ and an L2 decay of $10^{-5}$.
We use a prediction threshold of $0.3$ which was decided based upon a held out synthetically generated dataset.
All experiments are run on a Linux-based workstation made of an Intel Xeon W-2225 with 32 GB of memory and an NVIDIA GeForce RTX 3080 with 10 GB of memory.
We implemented our neural network in PyTorch and used a C++ implementation of MST~\cite{rmst_github}.

\paragraph{Synthetic nets}
The synthetic nets used for training are generated by sampling $N$ points (as pins) uniformly at random in the 2D plane, where $N \sim \operatorname{Uniform}(5,30)$.
We trained our model on $50$ million such nets.
Labels, i.e., the optimal set of Steiner points for these samples, were obtained using GeoSteiner~\cite{juhl2018geosteiner}.

\paragraph{Real nets}
We evaluate performance on real netlists from the ISPD 2005~\cite{ispd2005} and 2019~\cite{ispd2019} benchmarks.
These netlists are made from about 1,000 to 400,000 nets.
After filtering for nets with degree larger than $2$ and smaller than or equal to $64$, the median net degree is 4 and the mean degree is about 7. 
More statistics about these netlists are shown in Table~\ref{tab:datasets} in Appendix~\ref{sec:datasets}.

\paragraph{Baselines}
We benchmark \namemodel against several methods: GeoSteiner~\cite{juhl2018geosteiner} as an optimal solver, a plain Minimum Spanning Tree (MST) as the simplest approximation, Bi1S~\cite{kahng1992new} as a representative heuristic, and REST~\cite{liu2021rest} as the only neural alternative.
GeoSteiner gives a lower-bound on WL estimation---MST an upper-bound.
REST is a deep reinforcement learning method that comes in two variants: $T=1$ and $T=8$, where $T$ is the number of performed augmentations (i.e., $T=8$ applies rotations and flips to all nets), a hyper-parameter to be set at inference time.
$T$ trades off accuracy with cost: lower values of $T$ result in faster but less accurate predictions.
We use the source-code released by the authors of REST in PyTorch (Python), and standard implementations of Bi1S, GeoSteiner, and MST~\cite{rmst_github} in C++.

\subsection{Results}\label{sec:results} 

\begin{table}
    \begin{tabular}{@{} l rr rr rr @{}}
        \toprule
        \multirow{2}{0em}{$\frac{\text{model}\rightarrow}{\text{netlist}\downarrow}$} & MST & BI1S & \multicolumn{2}{c}{REST} & \multicolumn{2}{c}{\namemodel} \\
        \cmidrule(lr){4-5} \cmidrule(l){6-7}
        & & & $T=1$ & $T=8$ & small & large \\
        \midrule
        test1    & 7.298 & 0.067 & 0.423 & 0.108 & 0.715 & 0.503 \\
        test2    & 7.760 & 0.045 & 0.224 & 0.044 & 0.324 & 0.237 \\
        test3    & 7.940 & 0.047 & 0.298 & 0.026 & 0.237 & 0.132 \\
        test4    & 4.033 & 0.015 & 0.073 & 0.007 & 0.387 & 0.289 \\
        test5    & 5.913 & 0.028 & 0.247 & 0.027 & 0.777 & 0.557 \\
        test6    & 8.167 & 0.046 & 0.218 & 0.044 & 0.333 & 0.243 \\
        test7    & 7.937 & 0.050 & 0.224 & 0.044 & 0.330 & 0.239 \\
        test8    & 7.875 & 0.047 & 0.216 & 0.041 & 0.317 & 0.229 \\
        test9    & 7.942 & 0.048 & 0.223 & 0.044 & 0.328 & 0.237 \\
        test10   & 7.832 & 0.047 & 0.220 & 0.042 & 0.322 & 0.230 \\
        \addlinespace
        adaptec1 & 7.271 & 0.060 & 0.330 & 0.046 & 0.298 & 0.206 \\
        adaptec2 & 7.133 & 0.064 & 0.385 & 0.065 & 0.349 & 0.250 \\
        adaptec3 & 6.921 & 0.054 & 0.412 & 0.068 & 0.318 & 0.252 \\
        adaptec4 & 7.097 & 0.051 & 0.396 & 0.067 & 0.289 & 0.238 \\
        bigblue1 & 7.195 & 0.063 & 0.443 & 0.053 & 0.291 & 0.190 \\
        bigblue2 & 6.676 & 0.054 & 0.261 & 0.040 & 0.264 & 0.176 \\
        bigblue3 & 7.451 & 0.063 & 0.440 & 0.068 & 0.319 & 0.247 \\
        \addlinespace
        overall  & 7.430 & 0.052 & 0.294 & 0.050 & 0.317 & 0.232 \\
        \bottomrule
    \end{tabular}
    \caption{Wirelength (WL) estimation error (\%) against the optimal (GST).
        The reported error is the average across all nets, per netlist and overall.
        Lower is better.
    }\label{tab:all_quality}
\end{table}
\begin{table}
    \small
    \begin{tabular}{@{} l rrr rr rr @{}}
        \toprule
        \multirow{2}{0em}{$\frac{\text{model}\rightarrow}{\text{netlist}\downarrow}$} & GST & MST & BI1S & \multicolumn{2}{c}{REST} & \multicolumn{2}{c}{\namemodel} \\
        \cmidrule(lr){5-6} \cmidrule(l){7-8}
        & & & & $T=1$ & $T=8$ & small & large \\
        \midrule
        test1  &   0.93 & 0.02 & 2.05 & 98.01 & 111.59 & 1.00 & 0.55 \\
        test2  &  15.38 & 0.41 & 14.20 & 112.61 & 133.72 & 16.61 & 9.50 \\
        test3  &   0.37 & 0.03 & 1.43 & 83.71 & 94.66 & 0.81 & 0.47 \\
        test4  &   2.10 & 0.75 & 8.14 & 50.91 & 60.45 & 3.84 & 2.73 \\
        test5  &   0.41 & 0.05 & 1.71 & 81.30 & 92.09 & 0.93 & 0.57 \\
        test6  &  81.30 & 1.29 & 72.04 & 115.22 & 144.70 & 38.54 & 21.45 \\
        test7  & 162.16 & 3.16 & 150.72 & 117.99 & 158.01 & 77.18 & 44.74 \\
        test8  & 115.93 & 4.43 & 155.37 & 117.09 & 171.28 & 123.97 & 69.91 \\
        test9  & 415.69 & 7.50 & 324.28 & 116.06 & 189.92 & 207.50 & 111.91 \\
        test10 & 412.05 & 6.00 & 329.49 & 116.09 & 190.06 & 196.39 & 110.67 \\
        \addlinespace
        adaptec1 &  54.73 & 1.80 & 44.15 & 109.32 & 139.14 & 47.30 & 25.35 \\
        adaptec2 &  75.10 & 1.88 & 97.23 & 111.92 & 143.32 & 72.44 & 40.13 \\
        adaptec3 & 133.63 & 2.46 & 148.02 & 113.30 & 154.88 & 124.73 & 67.31 \\
        adaptec4 & 133.84 & 2.36 & 137.12 & 114.16 & 157.15 & 122.22 & 67.77 \\
        bigblue1 &  61.60 & 2.03 & 47.87 & 111.55 & 143.39 & 51.81 & 29.07 \\
        bigblue2 &  99.50 & 3.59 & 135.59 & 114.17 & 155.60 & 88.56 & 48.20 \\
        bigblue3 & 196.52 & 5.42 & 232.36 & 115.69 & 174.56 & 169.87 & 91.84 \\
        \addlinespace
        per net & 770 & 17 & 790 & 720 & 970 & 300 & 540 \\
        \bottomrule
    \end{tabular}
    \caption{
        Total runtime per netlist in seconds, and the mean per net in microseconds.
        Lower is better.
    }\label{tab:all_time}
\end{table}

\begin{figure*}
    \subfloat[Trading WL error (\%) for runtime by controlling the model capacity. These models were trained on synthetic nets, fine-tuned on \textit{adaptec4}, and evaluated on \textit{adaptec1}. \label{fig:cap_time_error}]{{
        \includegraphics[height=4.0cm]{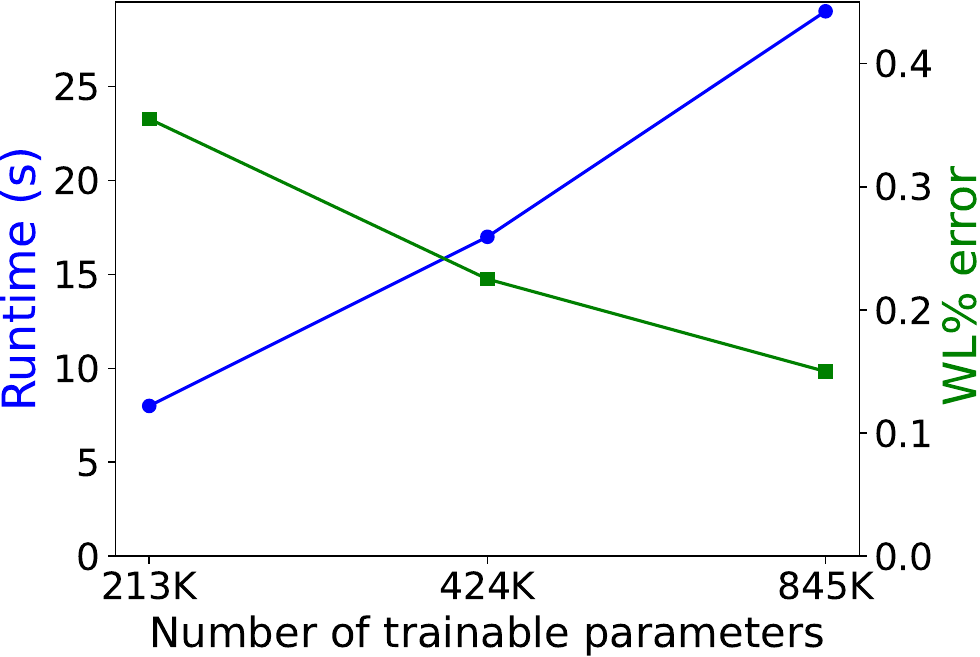}
    }}
    \hfill
    \subfloat[Training on more synthetic nets reduces WL error, at all model capacities. Evaluated on \textit{adaptec1}. \label{fig:more-data} ]{{
        \includegraphics[height=4.0cm]{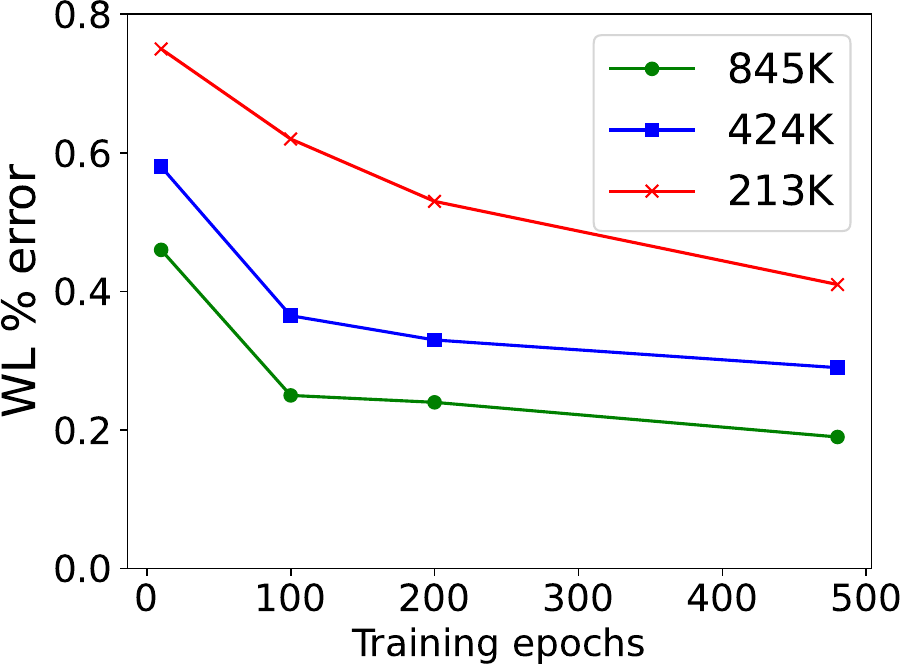}
    }}
    \hfill
    \subfloat[Thanks to batching, more GPU memory leads to lower runtime. Runtime reported for the 424K parameter model on \textit{adaptec1}. \label{fig:GPU_mem_time} ]{{
        \includegraphics[height=4.0cm]{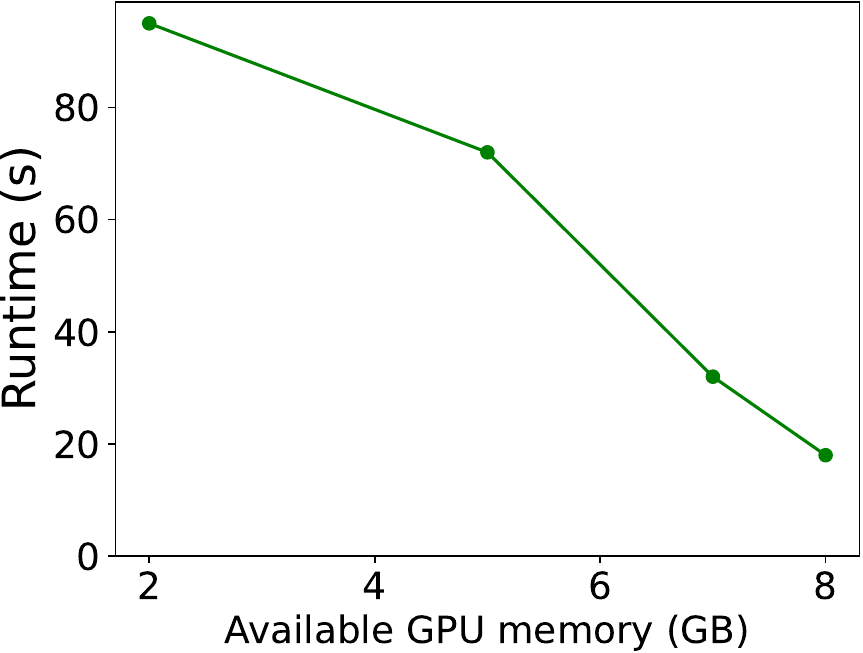}
    }}
    \vspace{-0.5cm}
    \caption{Scaling properties.
    }
\end{figure*}

\begin{table}
    \begin{tabular}{@{} c rr rr rr @{}}
        \toprule
        $\frac{\text{model}\rightarrow}{\text{degree}\downarrow}$ & MST & BI1S & \multicolumn{2}{c}{REST} & \multicolumn{2}{c}{\namemodel} \\
        \cmidrule(lr){4-5} \cmidrule(l){6-7}
        & & & $T=1$ & $T=8$ & small & large \\
        \midrule
          3--9 & 6.79 & 0.03 & 0.25 & 0.02 & 0.07 & 0.03 \\
        10--19 & 7.38 & 0.17 & 0.58 & 0.07 & 0.83 & 0.57 \\
        20--29 & 8.26 & 0.25 & 1.22 & 0.30 & 1.94 & 1.59 \\
        30--39 & 8.42 & 0.26 & 1.82 & 0.54 & 2.79 & 2.46 \\
        40--49 & 7.32 & 0.20 & 3.47 & 1.11 & 3.72 & 4.21 \\
        50--59 & 7.12 & 0.18 & 4.74 & 1.75 & 4.50 & 5.14 \\
        60--64 & 7.21 & 0.19 & 5.36 & 2.12 & 4.40 & 5.65 \\
        \bottomrule
    \end{tabular}
    \caption{Wirelength (WL) estimation error (\%), averaged over \textit{adaptec3} nets, grouped by degree.}\label{tab:quality_degree_wise}
\end{table}

\paragraph{WL estimation error} 
Table~\ref{tab:all_quality} shows the WL estimation error obtained by different methods against the minimal WL (given by GeoSteiner).
While \namemodel doesn't reach the performance of well-engineered heuristics, it improves on the faster variant of REST ($T=1$), the only neural alternative.
While REST with augmentations ($T=8$) achieves significantly better performance than our large model, it targets a corner of the cost--accuracy frontier.
However, for WL estimation in placement optimization, we expect errors within the $1\%$ range to be sufficiently low.
Hence we focused on achieving a fast model below the $1\%$ error mark.

Table~\ref{tab:quality_degree_wise} breaks down the estimation error per net degree.
We observe that the error of all methods grow with net degree.
Both \namemodel and REST with $T=8$ match Bi1S on nets of small degree, which dominate real netlists (as shown in Table~\ref{tab:datasets}).

\paragraph{Runtime}
Table~\ref{tab:all_time} shows the time required by different methods to estimate WL.
The duration of initial data processing is not included as it is shared amongst all methods.
For \namemodel, we report the total runtime: the forward pass of the neural model followed by the MST algorithm.
We note that the difference between the runtimes of REST that we report and the ones the authors report~\cite{liu2021rest} is due to potential differences in model loading and batching.
Namely, since REST uses different model checkpoints for different degrees, we evaluate different degree nets on different checkpoints, which have been provided for every $5$ degrees by the authors.
We then report the inference runtime of REST for sequentially processing batches of same-degree nets on the corresponding checkpoint for the closest multiple of $5$.
We observe that in most cases \namemodel is faster than REST.
We also observe a constant runtime overhead for REST, which we attribute to the sequential nature of degree-based batching and prediction.
On the other hand, for \namemodel, we divide nets into $5$ groups of similar degrees.
For each group, we use an optimized batch size, fitting as many nets as possible into a single forward pass.

\subsection{Fine-tuning on real nets}\label{sec:synthetic-real}

\begin{table}
    \begin{tabular}{@{} ll rrr @{}}
        \toprule
        model & test netlist & \multicolumn{3}{c}{training nets} \\
        \cmidrule(l){3-5}
        & & adaptec4 & synthetic & both  \\
        \midrule
        large & adaptec1 & 0.351 & 0.194 & 0.159 \\
        small & adaptec3 & 0.533 & 0.318 & 0.290 \\
        \bottomrule
    \end{tabular}
    \caption{
        Wirelength (WL) estimation error (\%) for different training strategies: training on \textit{adaptec4} only, training on synthetic nets only, or training on synthetic nets then fine-tuning on \textit{adaptec4}.
        Lower is better.
    }\label{tab:pre_ft_scratch}
\end{table}

In section \ref{sec:results}, we showed that training on synthetic nets achieves good results on real-world benchmarks without further fine-tuning, alleviating the need to train on scarce real designs.
If one nonetheless has access to relevant netlists---for example previous versions of the same design---training on them can improve accuracy~\cite{yue2022scalability} due to a lower discrepancy between the training and testing nets.

Table~\ref{tab:pre_ft_scratch} shows the WL estimation error for different training strategies.%
\footnote{
    These are not comparable to the errors reported in Table~\ref{tab:all_quality}.
}
Training on synthetic nets is better than training on a single real design.
With many parameters, our model has the capacity to overfit smaller datasets and to fail generalizing to different test datasets.
This emphasizes the importance of training on many synthetic nets (see also Figure~\ref{fig:more-data}).

However, training on synthetic then real nets, i.e., training then fine-tuning, shows a significant improvement (especially with the larger model).
This suggests that while synthetic training makes the model learn the fundamentals, fine-tuning enables tailoring to the real data distribution, without overfitting.
As a result, we expect training on cheap synthetic nets then fine-tuning on scarce real nets to further improve on the performance reported in Table~\ref{tab:all_quality}.

\subsection{Scaling properties}\label{sec:scaling}


\paragraph{Cost--accuracy tradeoff.}
Figure~\ref{fig:cap_time_error} shows the runtime and WL estimation error obtained by \namemodel for different model capacities (number of trainable parameters).
As the number of parameters increase, the error decreases and the runtime increases---allowing for an easy control of the cost--accuracy tradeoff.

\paragraph{More data}
Figure~\ref{fig:more-data} shows how training on more synthetic nets reduces the WL estimation error till convergence.

\paragraph{Parallelism}
As \namemodel, unlike REST, can process nets of different degree simultaneously, increasing GPU memory allows for more nets to be processed simultaneously, irrespective of a netlist's degree distribution.
Figure~\ref{fig:GPU_mem_time} shows that the runtime of \namemodel indeed decreases linearly as GPU memory increases.

\section{Conclusion}

We introduced \namemodel, a Graph Transformer neural model, to predict the Steiner points on a Hanan grid graph in \textit{one shot}.
Wirelength (WL) is then estimated through the construction of Rectilinear Steiner Minimum Trees (RSMTs). 


\paragraph{Discussion}
Although \namemodel achieves high performance at faster runtime when compared to GeoSteiner, we acknowledge that heuristics, such as SFP, for computing the RSMT can achieve competitive performance and prediction speed through efficient implementations.
This is common in the ML for combinatorial optimizaiton literature~\cite{joshi2020learning}.
However, algorithms, such as GeoSteiner, Bi1S, or SFP make discrete choices in order to generate Steiner points, impairing differentiability and often fixing a single tradeoff in terms of speed and accuracy.
In this work, we aim to advance the frontier of ML-based methods for approximating the RSMT.
We show that, compared to the state-of-the-art neural baseline REST, \namemodel is able to achieve a competitive solution quality while requiring only a single neural model.
This has potential advantages in terms of inference speed and memory requirements.
Furthermore, we show that \namemodel can be fine-tuned to a task-specific data distribution (Section~\ref{sec:synthetic-real}), allowing for further performance improvements when iterating through netlist revisions.



\paragraph{Future Work}
To simplify modeling, we treated each Steiner point independently during training and inference.
Since the Steiner point problem is inherently combinatorial, future work could explore joint or conditional Steiner points models to further enhance the quality of predictions.
Furthermore, our proposed method first predicts the Steiner points and then computes MST to obtain wirelength.
An alternative to this approach is to use a neural network to directly predict the wirelength of RSMT by treating it as a regression task.
This alternative promises faster execution time and an end-to-end differentiable path from wirelength to the input points.

\bibliography{references}
\bibliographystyle{ACM-Reference-Format}

\appendix

\section{Dataset description}\label{sec:datasets}


\begin{table}[h!]
    \begin{tabular}{@{} ll rrr @{}}
    \toprule
    dataset & netlist & \#nets & \multicolumn{2}{c}{degree} \\
    \cmidrule(l){4-5}
    & & & mean & median \\
    \midrule
    ISPD2019~\cite{ispd2019} & test1  &   1,199 & 10.93 & 4 \\
                             & test2  &  30,471 &  7.31 & 4 \\
                             & test3  &   3,498 &  5.49 & 4 \\
                             & test4  &  60,104 &  3.86 & 3 \\
                             & test5  &   5,467 &  5.89 & 4 \\
                             & test6  &  76,169 &  7.32 & 4 \\
                             & test7  & 152,171 &  7.33 & 4 \\
                             & test8  & 228,146 &  7.33 & 4 \\
                             & test9  & 380,151 &  7.34 & 4 \\
                             & test10 & 380,151 &  7.34 & 4 \\
    \addlinespace
    ISPD2005~\cite{ispd2005} & adaptec1 & 101,003 & 6.75 & 4 \\
                             & adaptec2 &  98,138 & 7.19 & 4 \\
                             & adaptec3 & 177,997 & 7.04 & 4 \\
                             & adaptec4 & 179,689 & 6.70 & 4 \\
                             & bigblue1 & 114,750 & 6.83 & 4 \\
                             & bigblue2 & 203,712 & 6.15 & 4 \\
                             & bigblue3 & 292,528 & 7.17 & 4 \\
    \bottomrule
    \end{tabular}
    \caption{
        Description of the netlists used for evaluation.
        Nets with degree smaller than $3$ and larger than $64$ are excluded.
    }\label{tab:datasets}
\end{table}



\end{document}